# GNTeam at 2018 n2c2: Feature-augmented BiLSTM-CRF for drug-related entity recognition in hospital discharge summaries

Maksim Belousov[1], Nikola Milosevic[1], Ghada Alfattni[1], Haifa Alrdahi[1], Goran Nenadic[1,2,3]

[1]School of Computer Science, The University of Manchester
[2]Health eResearch Center, The University of Manchester
[3]The Alan Turing Institute

## ABSTRACT

**Objective**

Monitoring the administration of drugs and adverse drug reactions are key parts of pharmacovigilance. In this paper, we explore the extraction of drug mentions and drug-related information (reason for taking a drug, route, frequency, dosage, strength, form, duration, and adverse events) from hospital discharge summaries through deep learning that relies on various representations for clinical named entity recognition.

**Materials and Methods**

This work was officially part of the 2018 n2c2 shared task, and we use the data supplied as part of the task. We developed two deep learning architecture based on recurrent neural networks and pre-trained language models. We also explore the effect of augmenting word representations with semantic features for clinical named entity recognition. Our feature-augmented BiLSTM-CRF model performed with F1-score of 92.67% and ranked 4th for entity extraction sub-task among submitted systems to n2c2 challenge.

**Results**

The recurrent neural networks that use the pre-trained domain-specific word embeddings and a CRF layer for label optimization perform drug, adverse event and related entities extraction with micro-averaged F1-score of over 91%. The augmentation of word vectors with semantic features extracted using available clinical NLP toolkits can further improve the performance.

**Conclusion**

Word embeddings that are pre-trained on a large unannotated corpus of relevant documents and further fine-tuned to the task perform rather well. However, the augmentation of word embeddings with semantic features can help improve the performance (primarily by boosting precision) of drug-related named entity recognition from electronic health records.

# INTRODUCTION

Almost half of the British population take prescribed medications, and many people take multiple drugs simultaneously. Over 20% of adults are administered with five or more drugs [1]. Drug-drug interactions and potential adverse drug events (ADEs) are not uncommon: ADEs caused by drug-drug interactions account for more than 30% of reported adverse drug reactions [2]. The quality of life and response to treatments are often affected by adverse drug reactions [3] and around 7% of hospital admissions are attributed to adverse drug events [2].

Many adverse drug events, especially those that occur infrequently, cannot be predicted by toxicological testing on animals and controlled clinical trials on humans [4]. This monitoring has to be extended beyond the period of drug testing and clinical trials. Hospital discharge summaries present a potentially valuable source of information for monitoring drug and treatment administration and associated adverse drug events. Discharge summaries typically outline the patient's complaint at admission, diagnostic findings, therapy administered, the patient's response to the therapy, and recommendations on discharge. However, they are formatted as unstructured textual documents, requiring clinical natural language processing (NLP) to extract information of interest from free-text summaries. Recognition of clinical entities from unstructured documents is still an active research area [5-9], with recent advances in neural network architectures for NLP, such as language models and recurrent neural networks demonstrating improved performances form many tasks.

To assess the current state of the art, the 2018 National NLP Clinical Challenge (n2c2) track 2 aimed to evaluate the entity recognition subtask focused on identifying drugs and related entities (reason, route, frequency, dosage, strength, form, duration, and ADE) from discharge summaries. In this paper, we examine the efficiency of a framework we developed for the task, including the effect of augmenting pre-trained domain word embeddings with semantic features in order to improve the performance of clinical named entities recognition. The effects of semantic feature augmentation were evaluated on randomly initialised embeddings as well as on embeddings pre-trained on MIMIC-III data. The proposed architecture has been officially evaluated as part of the n2c2 challenge.

# RELATED WORK

There have been several approaches proposed to extract drug names and drug attributes from free-text data. Several rule-based methods have been developed using semantic lexicons to extract drugs and related information from biomedical publications [10], social media [11, 12] and electronic health records [13-19]. Earlier machine-learning studies employed traditional approaches [10, 20-22] such as conditional random fields (CRFs) for extracting entities and support vector machine (SVM) for relation extraction between them. More recently, deep learning models, particularly the recurrent neural network (RNN) models, attracted much attention in the NLP community. The majority of methods use word embeddings, generated by training a language model on a large corpus of unannotated domain-specific documents [23, 24]. They could be utilised as initial word representations and then fine-tuned for a specific clinical entity recognition task [25-28], so that the final target model could benefit from large amounts of unannotated data, making task-specific learning more efficient. Feature augmentation is a method that aims to improve word representations learnt by a neural network by combining them with human-engineered features. Lee et al. [29] demonstrated that performance of their text de-identification method (i.e. recognition of personal identifiable entities in discharge summaries) is improved when word embeddings are concatenated with the features representations learnt with the feedforward neural network.

Several shared tasks for extraction of drug-related information have been organised. For example, the 3rd i2b2 shared task in 2009 focused on the identification of medication mentions in discharge summaries, along with associated attributes - dosages, modes of administration, frequency, duration, and reasons for prescriptions [14]. More recently, the Medication and Adverse Drug Events from Electronic Health Records (MADE) shared-task has been organised to detect medications and adverse drug events in electronic health records (EHRs) [30]. The target entities comprise drug name, dosage, route, duration, frequency, indication, ADE, and other signs and symptoms. In addition to training data, participants were provided with pre-trained word embedding trained using Wikipedia, de-identified Pittsburgh EHR and PubMed articles [31]. The top-performing systems used bidirectional long short-term memory (BiLSTM) with pre-trained word embeddings and a CRF layer for prediction. For example, Wunnava et al. [32] developed a three-layer neural network architecture, consisting of character-based BiLSTM, word-based BiLSTM and a CRF layer. They demonstrated that the integration of BiLSTM-CRF along with the character and word embeddings achieved excellent accuracy. Dandala et al. [33] developed a similar BiLSTM-CRF architecture but augmented concatenated character and word embeddings with part-of-speech embeddings. They observed the importance of features as well as pre-trained embeddings.

There are several open source clinical NLP toolkits that can be used to extract information from electronic health record and clinical narratives. Such information could also be used as features for augmentation of word embeddings. cTAKES [34], for example, offers several analysis engines for various NLP and specific clinical tasks, such as event identification, terminology mapping, uncertainty detection, temporal expressions identification and extraction drug attributes. Similarly, CLAMP [35] contains several components to facilitate building customised pipelines for diverse clinical applications.

## MATERIALS AND METHODS

### Task

The 2018 n2c2 task (track 2) focused on the extraction of several entity types: drugs, reason for taking a drug, route, frequency, dosage, strength, form, duration, and adverse events. Table 1 provides detailed descriptions and examples for the entity types. We note that both *ADE* and *Reason* denote conditions, signs and symptoms observed in the patient. The *Reason* class denote conditions, signs and symptoms for which the drug was administered, while *ADE* denotes unwanted signs and symptoms that happened as a consequence of an administered drug.

*Table 1: Descriptions of entity types in the shared task*

| Entity type | Description | Examples |
|---|---|---|
| **Drug** | The product name of the drug or its chemical substance name. | coumadin, vancomycin, aspirin, lasix, prednisone, o2, vitamin k, packed red blood cells |
| **Strength** | The amount of chemical substance of a drug in a given dosage. | 8.6 mg, 2.5 mg/3 ml (0.083%), 400 unit, 100 unit/ml, 5% (700 mg/patch) |
| **Form** | The form in which a drug should be taken. | Tablet, capsule, cream, tablet sustained release 24 hr |
| **Frequency** | The rate at which drug should be taken over a particular period of time. | daily, prn, q4h (every 4 hours) as needed, qid |
| **Route** | The way by which a drug should be taken or the location of absorbing a drug into the body. | po, iv, by mouth, inhalation, p.o., topical, nasal, injection |
| **Dosage** | The amount of a drug that a patient should take. | one (1), sliding scale, taper, 2 units, 30 ml, 100 unit/ml, |
| **Reason** | The indication or reason for drug administration. | pain, constipation, anxiety, nausea, wheezing, atrial fibrillation, pneumonia, hypotension |
| **ADE** | The development of unfavourable event due to a drug intake. | rash, thrombocytopenia, toxicity, diarrhea, altered mental status |
| **Duration** | The length of a period of time during which a drug should be taken, | for 7 days, for one week, 5 days, few days, prn, chronically, until his ciwa was less than 10 |

**Dataset**

The dataset that was used for training and evaluation was provided as a part of the n2c2 shared-task (track 2). The train set contained 303 labelled discharge letters with nine drug-related entities described in Table 1, and the test set included 202 documents. The documents were sampled as clinical care health records from MIMIC-III clinical care database [36]. MIMIC-III is a publicly-available database comprising de-identified health-related data associated with approximately sixty thousand admissions of patients who stayed in critical care units of the Beth Israel Deaconess Medical Center between 2001 and 2012. The frequency of mentions for each entity type, as well as the average number of mentions per document and the average number of tokens per mention can be seen in Table 2. While the majority of classes in the train set had over 4,000 mentions, we note that *Duration* and *ADE* had less than 1,000 instances each.

*Table 2: Number of mentions, average mention per document and average number of tokens per mention for each of the entity types in the train and test sets*

| Entity type | Mentions | | Avg. mention/document | | Avg. token/mention | |
|---|---|---|---|---|---|---|
| | train | test | train | test | train | test |
| **Drug** | 16,225 | 10,583 | 53.55 | 52.39 | 1.22 | 1.21 |
| **Strength** | 6,691 | 4,231 | 22.08 | 20.95 | 1.87 | 1.85 |
| **Form** | 6,651 | 4,358 | 21.95 | 21.57 | 1.67 | 1.66 |
| **Frequency** | 6,281 | 4,015 | 20.73 | 19.88 | 3.03 | 2.99 |
| **Route** | 5,467 | 3,513 | 18.07 | 17.39 | 1.07 | 1.07 |
| **Dosage** | 4,221 | 2,681 | 13.93 | 13.27 | 2.98 | 3.04 |
| **Reason** | 3,855 | 2,559 | 12.72 | 12.67 | 1.80 | 1.81 |
| **ADE** | 959 | 625 | 3.17 | 3.09 | 1.80 | 1.76 |
| **Duration** | 592 | 380 | 1.95 | 1.88 | 2.66 | 2.58 |

**Neural network architectures for clinical entity recognition**

We propose two deep learning architectures for recognising drug-related named entities in clinical texts. Initially, text segments were separated into word (token) sequences and all class mentions were converted into label sequences using the IOB (inside-outside-beginning) tagging format.

The word embedding layer transforms input raw words into vectors. The word representations were passed into the bidirectional recurrent neural network with long short-term memory units to learn important word-level features and transform them into the sequence label scores. The number of units for each RNN chain (i.e. backward and forward) was set to 70% of the input token representation size. The CRF layer is employed to optimize predictions across the whole sequence (i.e. text segment). Finally, the labels were combined into named entities by merging consecutively labelled B- or I- tags of the same class.

In order to explore the effect of adding semantic features, we have created a variation of the word-only architecture described above. We augment the word representations with semantic feature representations extracted using the CLAMP and cTAKES clinical pipelines. After processing discharge summaries through the pipelines, we extracted all token-level semantic tags (i.e. problem, treatment, test, temporal, negation, severity degree, body location, change, uncertainty) with associated assertion tag attributes (i.e. present or absent) from CLAMP and all semantic tags (i.e. Medication, DiseaseDisorder, SignSymptom, AnatomicalSite, Procedure) from cTAKES. We have merged cTAKES and CLAMP semantic tags to create a comprehensive set of features. For each pipeline, words are tagged with the corresponding semantic tag and attribute (if available) or using outside (i.e. O) tag otherwise. A direct mapping of semantic tags extracted using clinical pipelines to target entities is not feasible, because certain entities (such as Frequency or Route) are not presented among semantic tags, while other semantic tags (such as SignSymptom or DiseaseDisorder) are too broad. Thereby, the representations of the semantic tags could be learned simultaneously with word representations and concatenated together to form the final augmented token representations. The dimensionality of semantic representations has been set to 50. The architecture of the feature-augmented BiLSTM-CRF is presented in Figure 1.

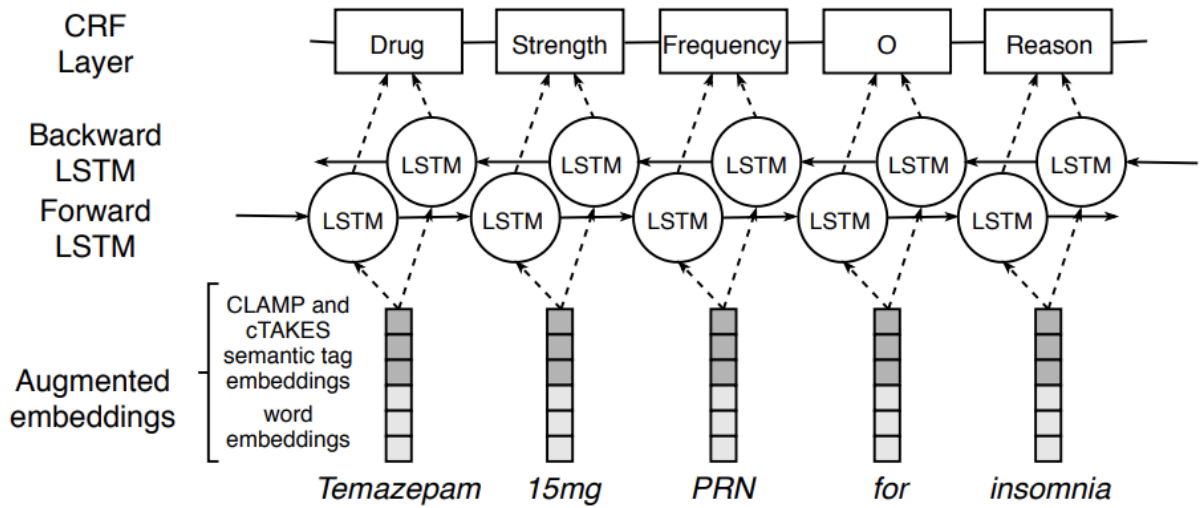

*Figure 1: The architecture of the feature-augmented bidirectional LSTM with the CRF (BiLSTM-CRF) label optimisation layer. Augmented embeddings consist of word embeddings and embeddings of semantic tags (extracted from CLAMP and cTAKES).*

**Experiments**

We have utilised the unstructured corpus of ~2 million discharge summaries from MIMIC-III dataset to learn 100-dimensional word embeddings using word2vec skip-gram model. The word2vec is a widely used method to generate word embeddings, supporting two different architecture types that determine how the context of the word is modelled [23]. The skip-gram architecture performs better for infrequent words, compared to the continuous bag-of-words (CBOW) model [37]. In order to investigate the effect of pre-trained embeddings for both word-only and feature-augmented BiLSTM-CRF architectures, we initialise the embedding layer weights with vectors from pre-trained word2vec embeddings and compared it with random vectors drawn from the uniform distribution [38]. To explore the efficiency of semantic features we performed feature-augmentation of both pre-trained and randomly initialised word vectors.

All four models were trained using the RMSProp [39] optimisation algorithm. We have reserved 10% of the data for validation and the number of epochs was determined by early stopping criteria (i.e. after 3 epochs with no improvement on the validation set).

**Evaluation methodology**

The primary evaluation metric for ranking systems in n2c2 challenge is the lenient micro-averaged F1-score. The strict metric counts only exact entity matches as correct, whereas the lenient metric does not take into account entity boundaries, considering all partial matches (overlapping entities) as correct. To evaluate our models, we performed hold-out cross-validation (using training and testing sets). We reported lenient precision, recall and F1-score.

**RESULT**

The results of the lenient evaluation of word-only and feature-augmented (with semantic features obtained from CLAMP and cTAKES) BiLSTM-CRF models with randomly initialised embeddings are presented in Table 3.

*Table 3: The evaluation results of word-only and feature-augmented BiLSTM-CRF models with randomly initialised embeddings on the test set (202 documents)*

| | Word-only BiLSTM-CRF (random) | | | Feature-augmented BiLSTM-CRF (random) | | |
|---|---|---|---|---|---|---|
| **Entity type** | **Precision** | **Recall** | **F1-score** | **Recall** | **Precision** | **F1-score** |
| **Strength** | 97.25 | 96.10 | 96.67 | 97.12 | 96.41 | **96.76** |
| **Frequency** | 96.83 | 95.07 | 95.94 | 96.81 | 94.60 | 95.69 |
| **Form** | 96.95 | 93.44 | 95.16 | 97.38 | 92.89 | 95.08 |
| **Route** | 97.19 | 92.63 | 94.85 | 96.60 | 93.14 | 94.84 |
| **Drug** | 95.81 | 90.16 | 92.90 | 96.12 | 90.61 | **93.28** |
| **Dosage** | 94.99 | 89.78 | 92.31 | 94.75 | 89.59 | 92.10 |
| **Duration** | 80.99 | 72.89 | 76.73 | 84.08 | 73.68 | **78.54** |
| **Reason** | 70.10 | 51.86 | 59.61 | 76.13 | 53.34 | **62.73** |
| **ADE** | 52.86 | 19.20 | 28.17 | 50.24 | 33.92 | **40.50** |
| **Overall (micro)** | 94.31 | 87.67 | 90.87 | 94.53 | 88.16 | **91.23** |
| **Overall (macro)** | 94.01 | 86.03 | 89.62 | 93.74 | 86.28 | **89.65** |

Overall, the performance of the feature augmented model was better for five classes (*Strength, Drug, Duration, Reason and ADE*) compared to the word-only model, and the decrease for the remaining four classes (*Frequency, Form, Route and Dosage*) was between 0.01-0.25. However, the overall micro-averaged and macro-averaged F1-scores increased by 0.35 and 0.03 respectively. Major increase was noticed for *Reason* and *ADE* classes (by 3.12 and 12.33 respectively), potentially due to adding relevant semantic tags (e.g. DiseaseDisorder, SignSymptom, problem) from clinical pipelines. We presented two confusion matrices calculated on the token level for word-only and feature-augmented BiLSTM-CRF models with randomly initialised embeddings in Figure 2. Each row of the confusion matrix is presenting the token instances of an actual class, whereas each column is presenting the token instances of a predicted class.

|  | **BiLSTM-CRF (random)** | | | | | | | | | **Feature-augmented BiLSTM-CRF (random)** | | | | | | | | |
| --- | --- | --- | --- | --- | --- | --- | --- | --- | --- | --- | --- | --- | --- | --- | --- | --- | --- | --- |
| Strength | **9675** | 14 | 17 | 0 | 21 | 69 | 0 | 4 | 3 | 311 | **9710** | 5 | 19 | 0 | 20 | 67 | 0 | 1 | 0 | 292 |
| Frequency | 8 | **11469** | 20 | 2 | 2 | 2 | 14 | 4 | 0 | 799 | 8 | **11395** | 25 | 2 | 4 | 3 | 11 | 8 | 0 | 864 |
| Form | 6 | 7 | **6866** | 46 | 48 | 46 | 0 | 3 | 0 | 242 | 10 | 11 | **6840** | 61 | 50 | 48 | 0 | 3 | 1 | 240 |
| Route | 6 | 14 | 44 | **3484** | 14 | 3 | 0 | 10 | 1 | 277 | 7 | 17 | 49 | **3514** | 11 | 7 | 0 | 4 | 0 | 244 |
| Drug | 18 | 40 | 13 | 2 | **11515** | 0 | 0 | 10 | 0 | 1458 | 18 | 11 | 18 | 4 | **11645** | 3 | 0 | 3 | 4 | 1350 |
| Dosage | 78 | 1 | 33 | 1 | 5 | **8067** | 3 | 6 | 0 | 393 | 110 | 0 | 28 | 1 | 4 | **8066** | 8 | 1 | 0 | 369 |
| Duration | 5 | 13 | 1 | 1 | 0 | 10 | **715** | 8 | 0 | 261 | 2 | 13 | 0 | 0 | 0 | 9 | **707** | 2 | 1 | 280 |
| Reason | 0 | 3 | 1 | 0 | 41 | 0 | 12 | **1852** | 23 | 2963 | 0 | 3 | 2 | 1 | 36 | 3 | 8 | **1888** | 76 | 2878 |
| ADE | 0 | 0 | 0 | 0 | 8 | 0 | 0 | 64 | **138** | 853 | 0 | 0 | 0 | 1 | 7 | 0 | 0 | 55 | **249** | 751 |
| O | 200 | 1945 | 428 | 64 | 654 | 132 | 120 | 729 | 118 | **92304** | 186 | 1908 | 401 | 78 | 639 | 134 | 113 | 592 | 203 | **92440** |
| | Strength | Frequency | Form | Route | Drug | Dosage | Duration | Reason | ADE | O | Strength | Frequency | Form | Route | Drug | Dosage | Duration | Reason | ADE | O |

*Figure 2: Token-level confusion matrix of word-only and feature-augmented BiLSTM-CRF with randomly initialised embeddings.*

It can be seen that both word-only and feature-augmented models were confused between specific entity pairs. Namely, *Dosage* was incorrectly labelled as *Strength* and vice-versa; *Form* class has been often misclassified as *Route*, *Drug* or *Dosage*; *ADE* was confused with *Reason*, while *Reason* was mislabelled as *Drug*. An example of confusion between *Strength* and *Dosage* can be seen in the phrase *"she received one litre of normal saline"*, where it is hard (even for a human) to distinguish if *"one litre"* is a strength or dosage. A potential approach to resolve this would be adding rules specific for some entity types or to use more semantic features.

Both *Reason* and *ADE* tokens were often not tagged as any entity class. In particular, over 60% of *Reason* tokens and over 80% of *ADE* tokens has been tagged as *Outside* class. Moreover, both entities usually describe diseases and symptoms, therefore there is a high confusion between them. The significant number of tokens (leading to about 6% of token-level recall loss) that were supposed to be tagged as *ADE* were misclassified as *Reason*. While the most confusing entity class for *Reason* was *Drug*. This potentially could happen because in some cases the indication (i.e. reason to take a drug) is included in the *Drug* entity (e.g. "pain medications", "anti-seizure medication"). By adding feature-augmentation, we observed an increase in the number of true positives for both ADE and Reason classes, in particular for ADE the increase was 80%. However, it had a negative impact on the number of true positive matches for *Frequency* class.

The results of the lenient evaluation of word-only and feature-augmented BiLSTM-CRF models with pre-trained word embeddings on MIMIC-III are presented in Table 4.

*Table 4: The evaluation results of word-only and feature-augmented BiLSTM-CRF models with pre-trained MIMIC-III embeddings on the test set (202 documents)*

| **Entity type** | **Word-only BiLSTM-CRF (MIMIC-III)** | | | **Feature-augmented BiLSTM-CRF (MIMIC-III)** | | |
|---|---|---|---|---|---|---|
| | **Precision** | **Recall** | **F1-score** | **Precision** | **Recall** | **F1-score** |
| **Strength** | 96.96 | 98.06 | 97.51 | 97.87 | 97.75 | **97.81** |
| **Frequency** | 96.97 | 95.86 | 96.42 | 96.66 | 96.06 | 96.36 |
| **Form** | 96.94 | 93.65 | 95.26 | 97.02 | 93.99 | **95.48** |
| **Route** | 95.66 | 94.22 | 94.94 | 96.47 | 94.22 | **95.33** |
| **Drug** | 95.06 | 94.96 | 95.01 | 96.26 | 94.28 | **95.26** |
| **Dosage** | 93.63 | 93.17 | 93.40 | 93.22 | 93.36 | 93.29 |
| **Duration** | 82.23 | 85.71 | 83.94 | 86.03 | 81.48 | 83.70 |
| **Reason** | 63.90 | 63.58 | 63.74 | 73.88 | 59.02 | **65.62** |
| **ADE** | 45.97 | 45.60 | 45.78 | 69.30 | 36.48 | **47.80** |
| **Overall (micro)** | 92.23 | 91.60 | 91.91 | 94.56 | 90.85 | **92.67** |
| **Overall (macro)** | 91.70 | 90.62 | 91.03 | 94.36 | 89.89 | **91.96** |

The performance increased for all entity types when embedding layer weights were initialised with vectors from MIMIC-III pre-trained word2vec embeddings. For the word-only model, the overall micro-averaged F1-score was 91.91 (compared to 90.87) and the feature-augmented model yielded 92.67 (compared to 91.23).

As noted in Table 4, the performance of the feature-augmented model was better for most of the classes compared to the word-only model, increasing both micro- and macro-averaged F1-scores by 0.76 and 0.93 respectively. Similarly, to randomly initialised models, for context-sensitive classes, such as *Reason* and *ADE* the F1-score increased by 1.88 and 2.02 respectively. However, for *Frequency*, *Dosage* and *Duration,* the performance dropped by less than 0.25.

In Figure 3 we presented two token-level confusion matrices for word-only and feature-augmented models with pre-trained word embeddings. We noticed that patterns in confusion between classes are similar to what we observed with randomly initialised models (Figure 2).

|  | \multicolumn{9}{c|}{BiLSTM-CRF (MIMIC)} | \multicolumn{9}{c|}{Feature-augmented BiLSTM-CRF (MIMIC)} |
|---|---|---|---|---|---|---|---|---|---|---|---|---|---|---|---|---|---|---|
|  | Strength | Frequency | Form | Route | Drug | Dosage | Duration | Reason | ADE | O | Strength | Frequency | Form | Route | Drug | Dosage | Duration | Reason | ADE | O |
| Strength | 9824 | 3 | 14 | 1 | 22 | 87 | 1 | 1 | 0 | 161 | 9764 | 4 | 15 | 1 | 18 | 90 | 1 | 1 | 0 | 220 |
| Frequency | 1 | 11547 | 25 | 3 | 8 | 2 | 17 | 5 | 0 | 712 | 1 | 11590 | 20 | 2 | 2 | 7 | 12 | 7 | 0 | 679 |
| Form | 9 | 7 | 6852 | 72 | 54 | 55 | 0 | 4 | 0 | 211 | 5 | 7 | 6887 | 55 | 64 | 54 | 0 | 3 | 0 | 189 |
| Route | 1 | 6 | 40 | 3557 | 11 | 2 | 1 | 12 | 0 | 223 | 1 | 8 | 52 | 3558 | 7 | 2 | 0 | 8 | 0 | 217 |
| Drug | 16 | 4 | 18 | 5 | 12194 | 1 | 0 | 17 | 4 | 797 | 8 | 3 | 12 | 3 | 12055 | 2 | 0 | 15 | 0 | 958 |
| Dosage | 94 | 1 | 35 | 2 | 3 | 8207 | 9 | 2 | 0 | 234 | 74 | 6 | 35 | 1 | 1 | 8190 | 4 | 3 | 0 | 273 |
| Duration | 3 | 10 | 0 | 0 | 2 | 5 | 819 | 5 | 2 | 168 | 0 | 14 | 0 | 1 | 0 | 10 | 796 | 2 | 2 | 189 |
| Reason | 0 | 2 | 1 | 1 | 42 | 0 | 3 | 2437 | 84 | 2325 | 0 | 1 | 1 | 1 | 25 | 0 | 3 | 2152 | 41 | 2671 |
| ADE | 0 | 0 | 0 | 0 | 7 | 0 | 0 | 98 | 364 | 594 | 0 | 0 | 0 | 0 | 6 | 0 | 0 | 82 | 260 | 715 |
| O | 214 | 1924 | 381 | 100 | 820 | 197 | 149 | 1574 | 477 | 190858 | 123 | 1930 | 400 | 93 | 605 | 175 | 110 | 733 | 96 | 192429 |

*Figure 3: Token-level confusion matrix for word-only and feature-augmented (using CLAMP and cTAKES semantic tags) BiLSTM-CRF with pre-trained word embeddings on MIMIC-III.*

As seen in Table 3 and 4, both word-only and feature-augmented architectures perform well achieving high precision, recall and F1-score (over 90%) for most of the classes except Duration, Reason and ADE. As noted in Table 2, Duration was the least frequent class in the dataset with only 592 instances. This potentially could have a negative impact on the performance, therefore, given more data, the performance would likely increase. We also observed from Table 3 and 4 – similarly to previous studies (e.g. [32]) – that *Reason* and *ADE* underperform due to their similarity (both denote conditions, signs and symptoms) – see confusion matrices (Figure 2 and 3). The difference is in the context in which they are mentioned, which can be challenging for a neural network to learn in particular as *ADE* was relatively infrequent.

The augmentation of word embeddings with semantic features demonstrated a slight increase in overall micro-averaged F1-score for clinical entity extraction. However, there were small differences between scores achieved by top-performing systems submitted to the n2c2 entity extraction task. The feature-augmented BiLSTM-CRF model with pre-trained word embeddings was ranked 4[th] among 30 teams. The system ranked 1[st] achieved micro-averaged overall F1-score of 94.18% (1.51% more than described system), while the system ranked as 10[th] yielded 91.4% (1.27% less than our system).

**CONCLUSION**

In this paper, we have described a recurrent neural network architecture that can be used for monitoring drug administration and potential adverse drug events in discharge summaries. In order to capture semantic regularities in the clinical language, word embeddings were pre-trained on a large unstructured corpus of clinical texts from the MIMIC-III database. Such embeddings have a comprehensive vocabulary which includes various synonyms and spelling variations of words. Therefore, the entity recognition model that utilise them is much less prone to overfitting and would require less labelled data to reach state-of-the-art performance. We have also examined the effects of augmenting word vectors with semantic features extracted using the CLAMP and cTAKES clinical pipelines. The evaluation results show that such augmentation generally improves the performance regardless of embedding initialisation method, especially for specific entity types such as ADE and Reason (for both lenient and strict scores was observed the same pattern of changes).

We also have noticed a few inconsistencies in the annotated dataset that might have contributed to the errors. One of the common errors appears when our model is annotating entities that are missing in the gold standard dataset. For example, the "three days" in the phrase "adding DRUG cover for the first three days of treatment" is not annotated as *Duration* in the gold standard while it seems to be an appropriate duration. There are also a few examples where the entities are annotated in various ways in the gold standard. For example, the word "injection" has been annotated as a *Form* in some cases and as a *Route* in others. Inconsistencies may also appear in annotation spans (e.g., annotating *a Dosage* or *Strength* and *Form* separately in some case and jointly in others).

Future work could involve the investigation of other types of features commonly used in clinical entity recognition, such as part-of-speech tags, regular expressions and external gazetteers. To address the issue related to confusion between *ADE* and *Reason* entities, ontologies containing information about common indications of drugs and their known adverse events could be utilised. Also, available historical patient-specific EHR metadata (i.e. data extracted from EHRs for a particular patient), such as medical conditions, previously prescribed drugs and experienced ADRs could add an extra layer of details that could boost the performance of entity recognition.

Since 2018, a number of novel contextualised word representation models were published, such as ELMo [40], GPT-2 [41], BERT [42], ERNIE [43]. These models utilize deeper architectures to represent both context and knowledge about entities more efficiently. Fine-tuning these language models for a given domain (i.e. discharge summaries) and task (i.e. clinical entity extraction) has a potential to further improve the entity extraction performance. However, pre-training or fine-tuning these models may be both resource and time intensive.

Moving further from entity recognition, relationships between such entities can be extracted to associate drugs with the correct context. Relationship extraction can be reduced to a classification task and similar architectures based on feature-augmented recurrent neural network can be applied (i.e. representing a context of relation as a sequence of words between participating entities and then instead of labelling each word in a sequence, the output will be aggregated to predict a relation type for the whole sequence). The combination of entity recognition and relation extraction can enable a full-scale end-to-end drug monitoring system and further statistical analysis of extracted information, potentially leading to reducing the cost and improving the quality of monitoring drug administration and adverse drug events.

**ACKNOWLEDGEMENT**

We would like to thank the organisers of n2c2 for providing annotated corpus, guidelines and evaluation script.

**FUNDING STATEMENT**

This research received no specific grant from any funding agency in the public, commercial or not-for-profit sectors.

**COMPETING INTERESTS STATEMENT**

Maksim Belousov, Nikola Milosevic, Ghada Alfattni, Haifa Alrdahi, and Goran Nenadic have no conflicts of interest that are directly relevant to the content of this study

**CONTRIBUTION STATEMENT**

M.B., N.M., G.A. and H.A. contributed to the design and implementation of the research, to the analysis of the results and to the writing of the manuscript. G.N. supervised the project.